\documentclass{article}
\usepackage{codefuse_tech_report}
\usepackage[colorlinks = true,
            linkcolor = blue,
            urlcolor  = blue,
            citecolor = blue,
            anchorcolor = blue]{hyperref}   
\usepackage{microtype}
\usepackage{xurl}
\usepackage{booktabs}
\usepackage{enumitem}
\usepackage{multicol}
\usepackage{CJKutf8}
\usepackage{siunitx}
\usepackage{floatflt}
\usepackage{graphicx}
\usepackage{wrapfig}
\usepackage{authblk}
\usepackage{lipsum}
\usepackage{algorithm}
\usepackage{algorithmicx}
\usepackage{algpseudocode}
\usepackage{subfigure}
\usepackage{multirow}
\usepackage{pifont}  
\usepackage{algorithm}
\usepackage{algpseudocode}
\usepackage{tabularx}
\usepackage{booktabs}   
\usepackage{graphicx}   
\usepackage[table]{xcolor} 
\algnewcommand{\LeftComment}[1]{\Statex \(\triangleright\) #1}

\usepackage{array}
\usepackage{amsmath}
\usepackage{amssymb}
\usepackage{mathtools}
\usepackage{amsthm}

\usepackage[capitalize,noabbrev]{cleveref}
\usepackage{adjustbox} 

\theoremstyle{plain}

\theoremstyle{definition}

\theoremstyle{remark}

\newcommand{\ie}{\textit{i.e.}}
\newcommand{\eg}{\textit{e.g.}}

\colmfinalcopy

\usepackage[utf8]{inputenc}
\usepackage[T1]{fontenc}
\usepackage{caption} 
\usepackage{adjustbox} 
\usepackage{fontawesome5}
\usepackage{url}
\usepackage{hyperref}
\usepackage{tcolorbox}
\usepackage{threeparttable} 
\tcbuselibrary{skins,breakable}

\tcbuselibrary{skins}

\usepackage{color}
\usepackage{xcolor}
\usepackage{soul} 

\definecolor{tred}{RGB}{251, 130, 132}
\definecolor{torange}{RGB}{247, 162, 116}
\definecolor{tyellow}{RGB}{251, 218, 140}
\definecolor{tgreen}{RGB}{127, 204, 181}
\definecolor{tblue}{RGB}{89, 177, 215}
\definecolor{insightblue}{RGB}{162, 210, 255}
\definecolor{questionred}{RGB}{255, 175, 204}

\title{\centering OpAgent: Operator Agent for Web Navigation}

\author{%
Yuyu Guo\thanks{Equal Contribution.}$^{\phantom{*}}$
~~Wenjie Yang$^{^{*}}$
~~Siyuan Yang$^{^{*}}$
~~Ziyang Liu$^{^{*}}$
~~Cheng Chen
~~Yuan Wei
\\
\vspace{-10pt}
\bf
~~Yun Hu
~~Yang Huang
~~Guoliang Hao
~~Dongsheng Yuan
~~Jianming Wang
~~Xin Chen
\\
\bf
~~Hang Yu\thanks{Correspondence to: Hang Yu \textless hyu.hugo@antgroup.com\textgreater.}
~~Lei Lei
~~Peng Di \\
\vspace{10pt}
Ant Group \\
\vspace{10pt}
\faGithub ~\url{https://github.com/codefuse-ai/OpAgent}\\
\hspace{-10pt}~~~~~~~~\includegraphics[width=1em,height=1em]{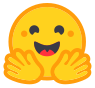} ~\url{https://huggingface.co/codefuse-ai/OpAgent}
}

\colmfinalcopy 

\begin{document}

\maketitle


\begin{abstract}
To fulfill user instructions, autonomous web agents must contend with the inherent complexity and volatile nature of real-world websites. Conventional paradigms predominantly rely on Supervised Fine-Tuning (SFT) or Offline Reinforcement Learning (RL) using static datasets. However, these methods suffer from severe distributional shifts, as offline trajectories fail to capture the stochastic state transitions and real-time feedback of unconstrained wide web environments. 
In this paper, we propose a robust Online Reinforcement Learning WebAgent, designed to optimize its policy through direct, iterative interactions with unconstrained wide websites. Our approach comprises three core innovations: 
1) Hierarchical Multi-Task Fine-tuning: We curate a comprehensive mixture of datasets categorized by functional primitives---Planning, Acting, and Grounding---establishing a Vision-Language Model (VLM) with strong instruction-following capabilities for Web GUI tasks. 
2) Online Agentic RL in the Wild: We develop an online interaction environment and fine-tune the VLM using a specialized RL pipeline. We introduce a Hybrid Reward Mechanism that combines a ground-truth-agnostic WebJudge for holistic outcome assessment with a Rule-based Decision Tree (RDT) for progress reward. This system effectively mitigates the credit assignment challenge in long-horizon navigation. Notably, our RL-enhanced model achieves a 38.1\% success rate (pass@5) on WebArena, outperforming all existing monolithic baselines. 
3) Operator Agent: We introduce a modular agentic framework, namely \textbf{OpAgent}, orchestrating a Planner, Grounder, Reflector, and Summarizer. This synergy enables robust error recovery and self-correction, elevating the agent's performance to a new State-of-the-Art (SOTA) success rate of \textbf{71.6\%}. 
\end{abstract}
\begin{figure}[h!]
    \centering
    \includegraphics[width=0.8\linewidth]{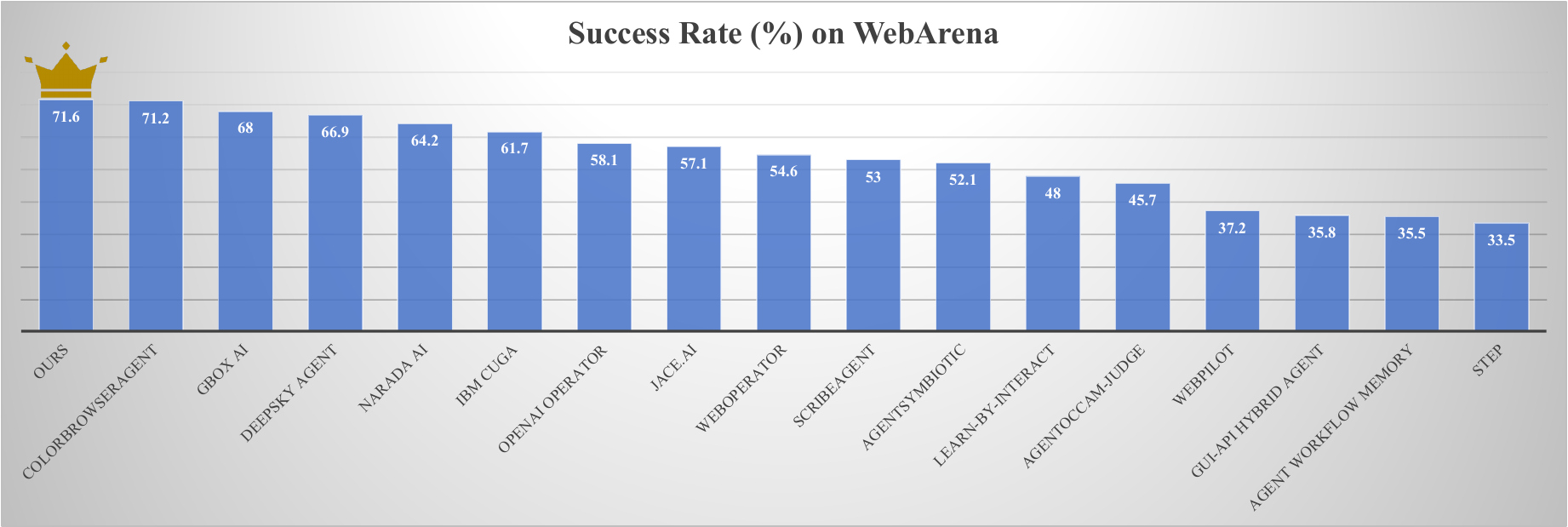}
    \caption{Our proposed OpAgent achieves a new state-of-the-art (SOTA) success rate of 71.6\% on the WebArena benchmark.}
    \label{fig:agentic_leadboard}
\end{figure}

\section{Introduction}

The rapid advancement of Large Language Models (LLMs)~\cite{llm:qwen3,llm:gpt4,llm:gemini25} and Vision-Language Models (VLMs)~\cite{mllm:qwen3vl,mllm:cogvlm,mllm:internvl35} has brought the development of autonomous agents—capable of navigating complex environments to fulfill user instructions—within practical reach. Previous research has extensively explored autonomous agents~\cite{agent:uitars,agent:cogagent} across various domains, spanning PC~\cite{agent:pc-tue,agent:pcagent}, mobile~\cite{agent:mobilerl,agent:mobileagentv3}, and web~\cite{agent:web-color,agent:web-tti} environments. 

Compared to desktop and mobile interfaces, the web presents three unique hurdles:
\begin{itemize}
    \item \textbf{Unstructured Complexity:} The underlying DOM tree is often orders of magnitude larger and more cluttered than desktop accessibility trees, laden with redundant or hidden metadata.
    \item \textbf{Temporal Volatility:} Websites exhibit extreme dynamicity, where asynchronous updates and ephemeral content render static, offline datasets rapidly obsolete.
    \item \textbf{Interaction Latency and Implicit Logic:} Many web operations require real-time feedback to verify success (\eg, hover-triggered menus), creating an interactive closed-loop that offline learning is inherently unable to simulate.
\end{itemize}

Despite these challenges, conventional paradigms predominantly rely on offline learning, which suffers from two fundamental limitations. 
Regarding \textit{unstructured complexity}, existing methods~\cite{bench:mind2web} heavily depend on text-based HTML or DOM tree parsing. Such representations are inherently fragile and sensitive to the immense noise found in modern websites, often overwhelming the agent's reasoning capacity. 
Regarding \textit{volatility and implicit logic}, previous ``static'' approaches~\cite{dataset:uground,dataset:webdreamer}—whether based on SFT or Offline RL—inevitably encounter severe \textbf{distributional shifts}. Since offline trajectories are pre-recorded, they fail to capture stochastic state transitions or provide the exploration-driven feedback loop necessary for mastering implicit interaction logic. Consequently, without the benefit of real-time trial-and-error, agents trained offline struggle to recover from execution failures or adapt to the fluid nature of live web environments. 

To bridge the gap between static training and dynamic execution, we propose \textbf{OpAgent}, a robust framework designed for autonomous web navigation. Our approach systematically tackles the aforementioned hurdles through three core strategies:

First, to navigate the \textbf{unstructured complexity} of the web, we transcend traditional text-centric paradigms. Instead, we leverage a Vision-Language Model (VLM)~\cite{mllm:qwen25vl,mllm:qwen3vl} to directly perceive visual signals, mimicking human-centric web interaction. This allows the agent to capitalize on spatial semantics and visual hierarchies that are more stable and informative than volatile HTML code. To forge a robust foundational policy, we curate a multi-task dataset categorized into three functional primitives: \textbf{Grounding, Planning, and Acting}, followed by a weighted multi-task Supervised Fine-Tuning (SFT)~\cite{multitask:Dynamic_Task,multitask:MFTCoder} process.

Second, to overcome \textbf{temporal instability} and \textbf{implicit logic}, we introduce an Online Agentic Reinforcement Learning framework. Unlike static training, our framework enables the agent to interact with live websites in real-time, effectively mitigating distributional shifts through autonomous exploration. We optimize the policy using a \textbf{Hybrid Reward Mechanism}, which synergizes a high-level VLM-based \textbf{WebJudge}~\cite{bench:online_mind2web} for holistic outcome verification with a \textbf{Rule-based Decision Tree (RDT)} for granular progress rewards.

Finally, we orchestrate these optimized capabilities within a modular architecture. By coordinating a specialized \textbf{Planner, Grounder, Reflector, and Summarizer}, \textbf{OpAgent} achieves sophisticated reasoning and robust self-correction, ensuring high success rates in long-horizon tasks. 

Our main contributions are summarized as follows: 
\begin{itemize}
    \item \textbf{Hierarchical Multi-Task Skill Acquisition:} We establish a foundational policy by curating a comprehensive dataset focused on \textit{Planning, Grounding, and Acting}. By prioritizing visual perception over brittle HTML parsing, our model effectively tames the unstructured complexity of modern web environments.
    \item \textbf{Agentic Online RL with Hybrid Rewards:} We develop an online RL pipeline that optimizes policies through live web interactions. By synergizing the \textbf{WebJudge} evaluator with a \textbf{Rule-based Decision Tree (RDT)}, we provide dense, ground-truth-agnostic supervision that effectively alleviates credit assignment challenges in the wild. Our RL-enhanced monolithic model achieves \textbf{38.1\% (pass@5)} on WebArena, outperforming existing monolithic baselines.
    \item \textbf{Collaborative Architecture and SOTA Performance:} We introduce a modular agentic framework that orchestrates specialized roles for error recovery. \textbf{OpAgent} achieves a new State-of-the-Art (SOTA) success rate of \textbf{71.6\%} on the WebArena benchmark, securing the top position on the leaderboard (January 2026).
\end{itemize}

\section{Related Work}
\label{sec:related_work}

\subsection{Autonomous Agents with LLMs and VLMs}
The emergence of Large Language Models (LLMs)~\cite{llm:gpt4,llm:qwen3} and their vision-language counterparts (VLMs)~\cite{mllm:qwen3vl,mllm:cogvlm} has revolutionized the development of autonomous agents. Early works focused on text-based reasoning and tool use in sandboxed environments~\cite{agent:react,agent:toolformer}. Recently, VLM-based agents have demonstrated superior performance in GUI-based tasks by directly perceiving visual signals~\cite{agent:uitars,agent:cogagent}. However, most of these agents are designed for general-purpose interaction and often lack the specialized policy optimization required for the highly dynamic and unstructured web environment.

\subsection{Web Navigation Agents}
Web navigation is a long-standing challenge in the AI community. Traditional methods predominantly relied on DOM tree parsing and text-centric representations~\cite{bench:mind2web}. While effective for simple layouts, they struggle with the noise and complexity of modern websites. To address this, recent works such as WebVoyager~\cite{bench:webvoyager} and SeeAct~\cite{agent:seeact} have transitioned to vision-based interaction. Despite their progress, these agents primarily rely on prompt engineering or Supervised Fine-Tuning (SFT) on static datasets. As highlighted in our study, these ``static'' paradigms suffer from significant distributional shifts when deployed in real-world, wild web environments. 

\subsection{Reinforcement Learning for Agents}
Reinforcement Learning (RL) has been widely explored to enhance agent decision-making. Initial attempts focused on Offline RL, where policies are optimized using pre-recorded expert trajectories~\cite{rl:guir1,rl:guig1}. However, offline methods are constrained by the quality and coverage of the static datasets. To bridge the gap between training and execution, recent research has begun exploring Online RL in simulated environments with text observation~\cite{rl:webr1}. Our work advances this frontier by implementing an Online Agentic RL framework in the wild with a vision-language model, utilizing a ground-truth-agnostic Hybrid Reward Mechanism to provide dense supervision without the need for oracle trajectories.

\subsection{Agentic Architectures and Self-Correction}
Complex, long-horizon tasks often require more than a single-step policy. Architectures such as Reflexion~\cite{agent:reflexion} have introduced self-correction mechanisms where agents reflect on their past errors. In the web domain, multi-agent orchestration has also shown promise in decomposing complex goals into sub-tasks~\cite{webagent:colorbrowser}. Our Collaborative Agentic Architecture builds upon these concepts by orchestrating a modular team of Planner, Grounder, Reflector, and Summarizer, specifically optimized for the robust error recovery needed in web navigation.

\section{Method}
\label{sec:method}
\begin{figure}[h!]
    \centering
    \includegraphics[width=0.8\linewidth]{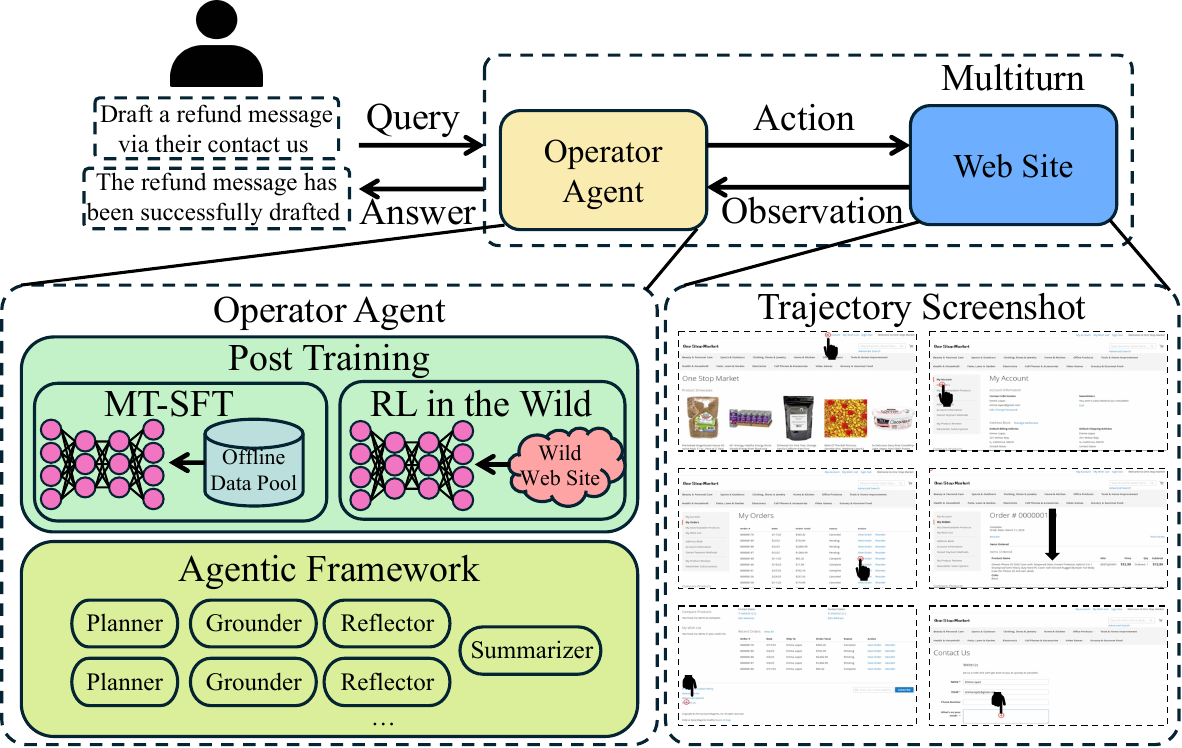}
    \caption{Overall architecture and training pipeline of O p A gen t. (Top) The system facilitates a multi-turn interaction loop where the Operator Agent executes actions and receives observations from live websites to fulfill user queries. (Bottom-Left) The development of the agent follows a hierarchical post-training paradigm: MT-SFT on offline data to establish foundational capabilities, followed by RL in the Wild for adaptive policy optimization in real-world environments. The agentic framework orchestrates modular roles including Planner, Grounder, Reflector and Summarizer. (Bottom-Right) A sample trajectory demonstrates the step-wise execution of a complex refund request task.}
    \label{fig:overview}
\end{figure}

As shown in \cref{fig:overview}, the OpAgent framework operates through a closed-loop interaction between the user, the Operator Agent, and the live web environment. Upon receiving a natural language instruction, the agent iteratively perceives the current visual state of the website, reasons over the multimodal information, and executes a sequence of actions until the task is successfully concluded with an answer.

To empower the agent with robust navigation in unconstrained environments, our training and execution paradigm is structured into three integrated components. First, we establish a cross-modal foundational policy through a \textbf{Hierarchical Multi-task Supervised Fine-tuning (MT-SFT)} stage, where the model learns fundamental interaction primitives including planning, grounding, and acting. Second, to bridge the gap between static data and dynamic web environments, we perform \textbf{Online Agentic RL in the Wild}, allowing the agent to refine its policy through real-time trial-and-error using a hybrid reward mechanism. Finally, during inference, these capabilities are orchestrated by an OpAgent architecture comprising specialized modules: Planner, Grounder, Reflector, and Summarizer. 
In the following sections, we elaborate on each component of our framework.

\subsection{Multitask Supervised Finetuning with Effective Weight}
\label{sec.msft}
\begin{figure}[h!]
    \centering
    \includegraphics[width=0.5\linewidth]{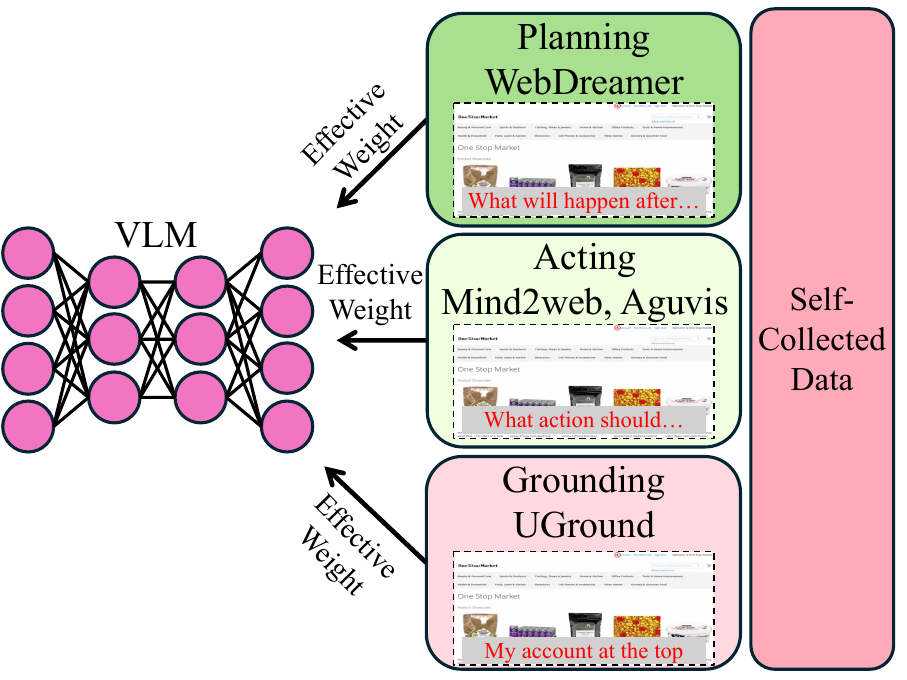}
    \caption{Illustration of the Hierarchical Multi-Task Supervised Fine-tuning (MT-SFT) pipeline. We initialize the VLM by joint training on a diverse mixture of self-collected data, categorized into three functional primitives: (1) \textbf{Planning} (via WebDreamer) for high-level goal decomposition and state prediction; (2) \textbf{Acting} (via Mind2Web and Aguvis) for low-level action execution; and (3) \textbf{Grounding} (via UGround) for spatial element localization. A \textit{Task-specific Effective Weighting} strategy is employed to balance the learning gradients across these heterogeneous tasks, ensuring a robust foundational policy for subsequent RL optimization.}
    \label{fig:msft}
\end{figure}
We categorize the fundamental capabilities of a web agent into three dimensions: \textbf{Planning}, \textbf{Acting}, and \textbf{Grounding}. Specifically, \textbf{Planning} involves assessing the functional affordance of UI controls, \ie, anticipating the subsequent page transitions and state changes triggered by a specific interaction. \textbf{Acting} entails determining the requisite operation to be performed on the current page to advance the task. Finally, \textbf{Grounding} addresses the spatial localization, specifying the precise coordinates or UI elements where the action should be executed. We foster the development of each functional primitive by incorporating targeted open-source datasets. For the \textit{Planning} dimension, we employ WebDreamer~\cite{dataset:webdreamer} to enhance the model's reasoning about state transitions. For Acting, we rely on the large-scale interaction trajectories from Mind2Web~\cite{bench:mind2web} and Aguvis~\cite{dataset:aguvis}. Finally, UGround~\cite{dataset:uground} is utilized to refine the agent's \textit{Grounding} precision through its high-quality spatial annotations. However, a direct concatenation of these heterogeneous datasets poses a significant data imbalance challenge. For instance, while Mind2Web contains only a few thousand steps, UGround reaches the million-scale. Consequently, naive joint training would inevitably cause the model to be dominated by the task with the larger data volume, compromising performance on smaller but crucial tasks. To mitigate this, we introduce a weighting strategy based on the effective number of samples~\cite{loss:cbloss}. Let $C$ be the number of different datasets and $n_i$ be the number of samples for dataset $i \in \{1, \dots, C\}$. We define the hyperparameter $\beta$ as:
\begin{equation}
    \beta = 1 - 10^{-k},
\end{equation}
where $k$ is a scaling factor obtained from the experimental configuration. The effective number of samples $E_{n_i}$ for dataset $i$ is calculated as:
\begin{equation}
    E_{n_i} = \frac{1 - \beta^{n_i}}{1 - \beta}.
\end{equation}
To balance the contribution of each task while maintaining gradient stability, we calculate the normalized \textbf{effective weight} $\alpha_i$:
\begin{equation}
    \alpha_i = C \cdot \frac{E_{n_i}^{-1}}{\sum_{j=1}^{C} E_{n_j}^{-1}} = C \cdot \frac{(1 - \beta^{n_i})^{-1}}{\sum_{j=1}^{C} (1 - \beta^{n_j})^{-1}}.
\end{equation}
During the MT-SFT stage, for a batch of size $B$, the total loss $\mathcal{L}_{SFT}$ is defined as the weighted average of individual sample losses:
\begin{equation}
    \mathcal{L}_{SFT} = \frac{1}{B} \sum_{m=1}^{B} \alpha_{dataset(m)} \cdot \left( \frac{1}{T_m} \sum_{t=1}^{T_m} \ell_{m,t} \right),
\end{equation}
where $\ell_{m,t}$ represents the cross-entropy loss for the $t$-th token in the $m$-th sample, $T_m$ is the sequence length, and $\alpha_{dataset(m)}$ is the effective weight corresponding to the dataset category of sample $m$. This mechanism ensures that the model develops balanced proficiency across Planning, Acting, and Grounding regardless of their original data scales.

\subsection{Online Agentic RL in the Wild}
\label{sec.wildrl}
Owing to the lack of stable and efficient reinforcement learning (RL) frameworks for web agents, we first developed a specialized browser-based RL infrastructure. We then optimized the agent’s policy through a novel \textbf{hybrid reward mechanism} that balances outcome-based and process-based supervision.
\subsubsection{Hierarchical Infrastructure for Web RL in the Wild}
\begin{figure}[h!]
    \centering
    \includegraphics[width=0.4\linewidth]{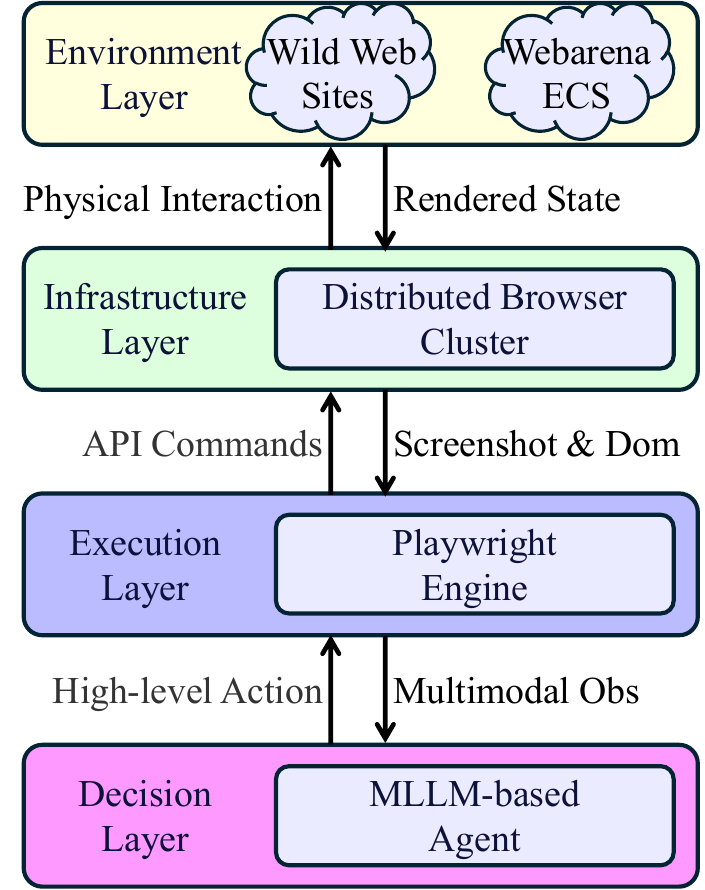}
    \caption{Hierarchical Infrastructure for the Web Agent RL. The system is organized into four functional layers: (1) the \textbf{Environment Layer} featuring a hybrid sandbox consisting of self-hosted the open \textbf{Wild Web} and \textbf{WebArena} on \textbf{Alibaba Cloud ECS}; (2) the \textbf{Infrastructure Layer} managing a distributed browser cluster for scalable data collection; (3) the \textbf{Execution Layer} utilizing a high-concurrency Playwright engine to translate semantic actions into API commands; and (4) the \textbf{Decision Layer} where the VLM-based agent performs reasoning and action generation. The solid arrows (left) denote the upward \textit{Action Flow}, while the dashed arrows (right) represent the downward \textit{Observation Flow} of multimodal feedback.}
    \label{fig:rl_inf}
\end{figure}
During the Online RL process, the model interacts with the environment throughout the rollout phase, following the dual-stream workflow illustrated in Figure~\ref{fig:rl_inf}. 
In the action flow, the VLM receives multimodal observations and generates high-level semantic actions in JSON format. Playwright then parses these JSON descriptions into executable API commands and dispatches them to the Distributed Browser Cluster. Within this infrastructure layer, individual browser instances perform the physical interactions on the target websites, including Wild Web sites and WebArena on ECS.
Subsequently, in the observation flow, the Browser Cluster renders the updated web states and captures raw data, including screenshots and DOM trees. Playwright then retrieves this raw information, performs data filtering and structuring, and returns a refined multimodal observation to the VLM. This four-tier loop ensures that the agent can collect high-quality trajectories at scale for iterative policy optimization. Furthermore, it is noteworthy that because the Playwright library is not inherently thread-safe, we implemented a dedicated thread pool during the initialization phase to enable asynchronous concurrency. This design effectively avoids the performance overhead of frequent instantiation and destruction of Playwright instances and mitigates potential memory leaks associated with repeated process creation and termination.

\subsubsection{Online Agentic RL with Hybrid Reward}
\begin{figure}[h!]
    \centering
    \includegraphics[width=0.8\linewidth]{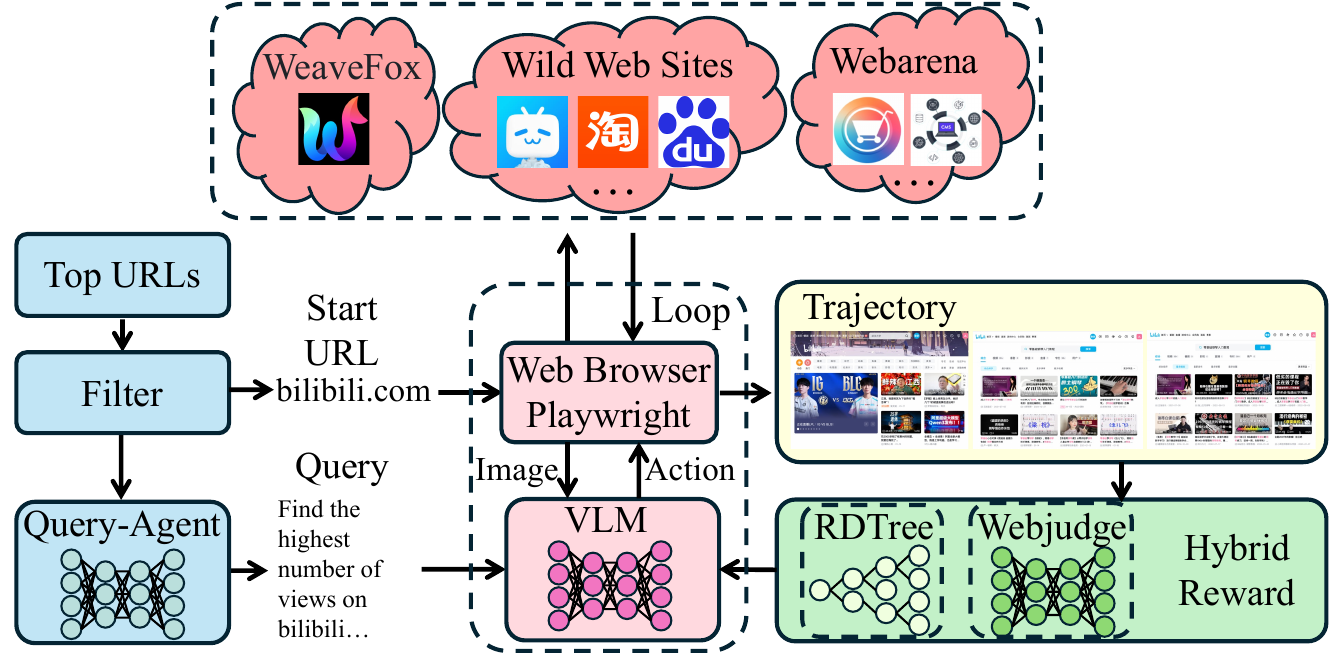}
    \caption{Overview of the OpAgent Training Infrastructure and Reinforcement Learning Loop. The framework consists of three core phases: (1) \textbf{Task Generation}, where a Query-Agent synthesizes realistic navigation goals on filtered top URLs; (2) \textbf{Interactive Rollout}, where the VLM-based agent interacts with a hybrid environment (including WeaveFox\protect\footnotemark, unconstrained Wild Web sites, and the self-hosted WebArena) via a high-concurrency Playwright engine; and (3) \textbf{Hybrid Reward Evaluation}. The reward system integrates an \textbf{RDTree} (Rule-based Decision Tree) to derive process-based rewards for intermediate steps, and \textbf{Webjudge}, which assesses visual trajectory screenshots to provide a holistic success score.}
    \label{fig:rl_hybrid}
\end{figure}
\footnotetext{\url{https://github.com/weavefox}}

To facilitate reinforcement learning in unconstrained environments, we first curate a collection of top-tier URLs across various domains and leverage a Large Language Model (LLM) to synthesize a diverse set of realistic user queries. During the rollout phase, the process begins with the Playwright engine navigating to the designated starting URL and transmitting the initial visual state to the Vision-Language Model (VLM). The agent then interacts with the website based on the generated query, as shown in Figure~\ref{fig:rl_hybrid}.
Notably, for multi-turn interactions, we adopt a text-centric history strategy: only the current visual observation (screenshot) is provided to the model at each timestep. While the history of previous reasoning and actions is preserved in textual format, all prior images are discarded. This design choice is informed by our empirical findings that current VLMs are prone to visual hallucinations when processing sequences of multiple images, which can significantly degrade the accuracy of decision-making.
The rollout process for each trajectory continues until the model signals task completion or reaches a predefined maximum step limit, after which the collected sequence is sent for reward evaluation. We employ Group Relative Policy Optimization (GRPO)~\cite{rl:deepseekr1,rl:grpo_math} to optimize the model. To prevent the issue of entropy collapse, we implemented the \textit{KL Cov} strategy~\cite{rl:entropy}. Unlike traditional global KL regularization, this approach selectively applies KL divergence constraints only to tokens that exhibit a high covariance between their advantage values and logits, thereby maintaining policy diversity during training.

To provide comprehensive supervision for policy optimization, we develop a multi-faceted reward function comprising structural and quality-based components. First, a Format Reward is employed to enforce structural compliance, ensuring the presence of a reasoning process (Chain-of-Thought) and the syntactic correctness of the output format.
Beyond basic formatting, we evaluate the quality of the navigation trajectory from two distinct perspectives: \ie, Outcome-based Evaluation and Process-based Supervision.

\begin{center} 
\begin{minipage}{0.6\linewidth} 
\begin{algorithm}[H]
\caption{Rule-based Decision Tree for Process Reward}
\label{alg:rdtree}
\begin{algorithmic}[1]
\Require Action $a_t$, Previous State $s_{t-1}$, Current State $s_t$, User Query $q$
\Ensure Step Reward $r_t$
\If{Execution Failed ($a_t$)}
\State \Return $r_{penalty}$
\EndIf
\If{URL Changed ($s_{t-1}, s_t$)} \Comment{Significant state transition}
\State \Return $0$
\ElsIf{Coordinates on Interactable Element ($a_t$)} \Comment{Valid UI affordance}
\State \Return $0$
\ElsIf{No Visual Change ($SSIM(s_{t-1}, s_t) == 1.0$)} \Comment{Redundant action}
\State \Return $r_{penalty}$
\ElsIf{VLM Judges Progress ($s_{t-1}, s_t, q$)} \Comment{Semantic verification}
\State \Return $0$
\Else
\State \Return $r_{penalty}$ \Comment{Visual noise or irrelevant change}
\EndIf
\end{algorithmic}
\end{algorithm}
\end{minipage}
\end{center}

\textbf{Outcome-based Evaluation (Webjudge): } We utilize Webjudge~\cite{bench:online_mind2web} as a global evaluator to assess task completion. By analyzing the screenshots captured throughout the entire trajectory, Webjudge assigns an outcome reward based on the final visual state and goal fulfillment. 
Specifically, Webjudge evaluates the quality of each trajectory across three primary dimensions:
\textbf{Task Completion (Score: -1--5):} This metric assesses the final fulfillment of the user's goal based on a result-oriented rubric. It quantifies whether the agent has successfully resolved the query through its interaction sequence. A score of $-1$ is specifically reserved for trajectories where the agent encounters insurmountable external obstacles, such as mandatory authentication (login walls) or network connectivity failures. To ensure the purity of the training signal and prevent the model from learning from environmental contingencies, we systematically discard all trajectories with negative scores during the RL training phase. \textbf{Action Validity (Score: 1--5):} This score evaluates the precision of element localization. By analyzing the visual variations between consecutive screenshots, Webjudge infers whether the agent's actions correctly targeted the intended UI components. \textbf{Trajectory Efficiency (Score: 1--5):} This metric measures the conciseness and logical flow of the interaction path, penalizing redundant steps or circular navigation according to predefined trajectory efficiency standards.

\textbf{Process-based Supervision (Rule-based Decision Tree): } To provide more granular feedback, we implement a process reward that evaluates the functional validity of each step. This mechanism verifies whether the agent's predicted actions (\eg, clicks or types) are executed on effective UI elements with actual affordance, thereby penalizing redundant or invalid interactions. The decision progress is shown in Algorithm~\ref{alg:rdtree}. 

The decision tree mechanism evaluates the validity of an action $a_t$ at state $s_t$ through the following hierarchical checks:
Navigation Verification (URL Change):
First, the system detects if the action triggered a page transition by comparing the current URL with the previous one. A change in URL implies a successful navigation event, which is considered a valid step (Reward $= 0$).
Affordance Validation (Element Interaction):
If the URL remains unchanged, the system verifies if the action (specifically for clicks or hovers) was performed on a valid UI element. It checks whether the action's coordinates $(x, y)$ fall within the bounding box of any interactable element (e.g., buttons, links, inputs) registered in the browser's accessibility tree. Valid interactions are deemed effective (Reward $= 0$).
Redundancy Check (Visual Stagnation):
For actions that neither change the URL nor hit a known interactable element, the system checks for visual state changes using the Structural Similarity Index (SSIM). If $SSIM(s_t, s_{t+1}) = 1.0$, indicating the screen is pixel-perfect identical to the previous frame, the action is classified as redundant or invalid (Reward $= \text{penalty}$, e.g., $-0.001$).
Semantic Progress Evaluation (VLM-based Check):
If the screen did change visually but failed the previous explicit checks, a generic Visual Language Model (VLM), \eg Qwen-VL-Max, is invoked as a fallback. The VLM compares the pre-action and post-action screenshots given the user's instruction to determine if the visual change represents meaningful progress. If the VLM confirms progress, the step is valid; otherwise, it is penalized.

\subsection{Operator Agentic Architecture for Web}
\label{sec.agenticarc}
\begin{figure}[h!]
    \centering
    \includegraphics[width=0.9\linewidth]{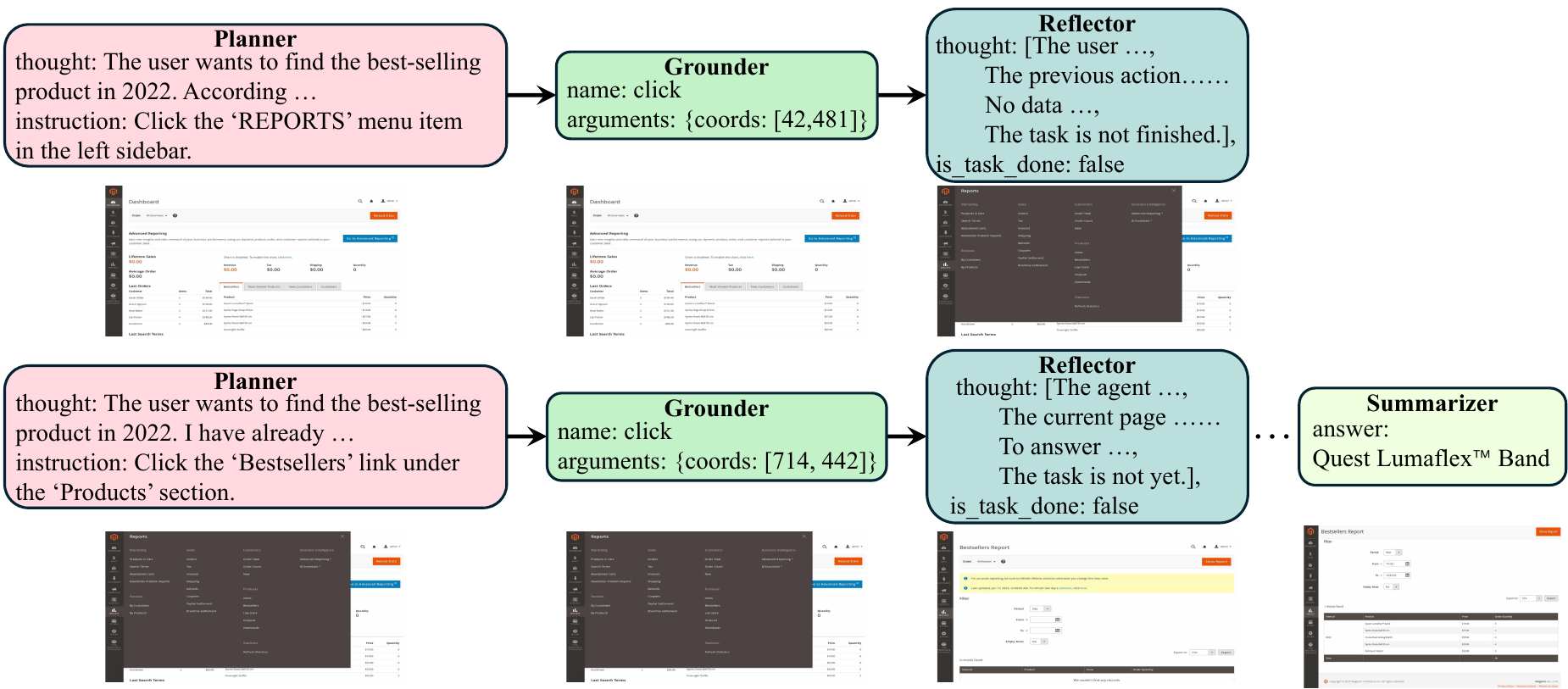}
    \caption{The Iterative Reasoning Pipeline of OpAgent. Our agentic framework decouples complex web navigation into four specialized modules: (1) the \textbf{Planner} generates high-level strategic thoughts and instructions; (2) the \textbf{Grounder} translates semantic instructions into precise executable actions and coordinates; (3) the \textbf{Reflector} evaluates the state transition and monitors task progress; and (4) the \textbf{Summarizer} module synthesizes the final answer once the goal is achieved. This loop ensures robust error correction and precise UI interaction.}
    \label{fig:agentic_frame}
\end{figure}

Due to the inherent complexity of web-based tasks, which often involve dozens of interaction steps, standalone models struggle to achieve satisfactory performance in isolation. At present, an agentic framework equipped with self-reflection and hierarchical task decomposition capabilities remains indispensable for navigating such intricate and long-horizon web environments. Accordingly, our work introduces a sophisticated agentic framework comprising four specialized modules: the Planner, the Grounder, the Reflector, and the Summarizer. The overall pipeline and the specific capabilities of our agentic framework are illustrated in Figure~\ref{fig:agentic_frame} and Table~\ref{tab:framework_modules}, respectively.

\paragraph{Planner: Strategic Decomposition} 
The Planner acts as the strategic core, responsible for decomposing the global \textit{user\_query} into atomic, executable steps. It analyzes the current visual state, the pending \textit{todo-list}, and feedback from the Reflector to synthesize a high-level semantic \textit{instruction}. Its key capabilities include:
\begin{itemize}
    \item \textbf{Context-Aware Planning:} Integrating historical trajectories and expert "tips" to dynamically adjust navigation strategies.
    \item \textbf{Adaptive Re-planning:} Utilizing reflection signals to correct erroneous paths or bypass failed interaction attempts.
\end{itemize}

\paragraph{Grounder: Visual-Action Mapping} 
The Grounder serves as the execution bridge that translates semantic intent into physical interactions. It receives the Planner's instructions and maps them to precise browser coordinates or tool-call parameters. Its key capabilities include:
\begin{itemize}
    \item \textbf{Visual Grounding:} Identifying specific UI elements (\eg, buttons, input fields) from raw screenshots.
    \item \textbf{Precise Actuation:} Generating tool calls (\eg, click, type, scroll) formatted for the underlying browser engine.
\end{itemize}

\paragraph{Reflector: Introspective Monitoring} 
The Reflector is an introspective module that ensures the reliability of the perception-action loop. It scrutinizes state transitions after each action to verify success and manage information extraction. Its key capabilities include:
\begin{itemize}
    \item \textbf{Factual Verification:} Confirming if an action achieved its intended effect based strictly on visual evidence, thus preventing hallucination-driven progress.
    \item \textbf{Incremental Extraction:} Identifying and recording goal-relevant information into structured \textit{marked notes}.
    \item \textbf{Blocker Detection:} Monitoring for "hard blockers" such as login walls or CAPTCHAs and triggering early termination when necessary.
\end{itemize}

\paragraph{Summarizer: Evidence Synthesis} 
The Summarizer performs a holistic review of the entire interaction episode. Once the Reflector signals completion or the step limit is reached, it distills the trajectory into a concise final answer. Its key capabilities include:
\begin{itemize}
    \item \textbf{Temporal Fusion:} Synthesizing information across the full sequence of screenshots and collected notes.
    \item \textbf{Goal Assessment:} Providing a final determination of task success and outputting the synthesized result to the user.
\end{itemize}

\begin{table}[ht]
\centering
\caption{Roles and Responsibilities within the Modular Agentic Framework.}
\label{tab:framework_modules}
\small
\begin{tabular}{@{}lll@{}}
\toprule
\textbf{Module} & \textbf{Core Responsibility} & \textbf{Key Output} \\ \midrule
\textbf{Planner} & Task decomposition and strategy formulation & Semantic Instruction \\
\textbf{Grounder} & Mapping semantic intent to UI coordinates & Tool Call \\
\textbf{Reflector} & Verifying action success and extracting data & Reflection Signal \& Notes \\
\textbf{Summarizer} & Final synthesis of trajectory evidence & Consolidated Answer \\ \bottomrule
\end{tabular}
\end{table}

\section{Experiments}
In our experiments, we utilize Qwen2.5-VL-72B-Instruction and Qwen3-VL-32B-Thinking as the primary models for post-training. In OpAgent Architecture, certain sub-agents are further empowered by Gemini-3-Pro to ensure robust performance across diverse web tasks on the WebArena benchmark. 

\subsection{Multitask Supervised Finetuning with Effective Weight}
For the Multi-task Supervised Fine-Tuning (MFT) with Effective Weight, we evaluate our approach on several established offline GUI benchmarks, including operation benchmarks (GUIAct~\cite{bench:guiact}) and grounding benchmarks (ScreenSpot~\cite{bench:screenspot} and ScreenSpot-v2~\cite{bench:screenspotv2}). The results are shown in Table~\ref{tab:guiact} and Table~\ref{tab:screenspot_combined}, as seen, our MFT model achieves competitive performance on these benchmark. To be more specific, our model demonstrates superior performance on text-based UI elements, achieving scores of 97.8 on PC text and 94.3 on Web text in the Screenspot V2 benchmark. This stems from our extensive collection of web data, as web environments feature a significantly higher proportion of text-containing controls compared to PC and mobile platforms.

\begin{table}[h!]
\centering
\caption{Performance comparison on Web-Multi subset of GUIAct benchmark} 

\label{tab:guiact}

\begin{tabular}{lcccc} 
\toprule
Method & Backbone & Type EM & Cli.Acc & StepSR  \\ \midrule
MiniCPM-GUI~\cite{bench:guiact}& MiniCPM-V &67.0 &45.5 & 47.5 \\
Qwen-GUI~\cite{bench:guiact}& Qwen-VL &68.9 &52.5 &46.8 \\
MFT (Ours) &Qwen2.5-VL-7B & 83.3& 64.4& 71.9\\
MFT (Ours) &Qwen2.5-VL-72B &\textbf{84.1} &\textbf{67.3} & \textbf{73.6}\\
\bottomrule
\end{tabular}
\end{table}

\begin{table*}[t]
\centering
\caption{Performance comparison on Screenspot (v1) and Screenspot-v2 (v2).} 
\label{tab:screenspot_combined}
\setlength{\tabcolsep}{5pt}
\resizebox{\textwidth}{!}{
\begin{tabular}{l cc cc cc cc cc cc cc} 
\toprule
& \multicolumn{2}{c}{Mobile-Text} & \multicolumn{2}{c}{Mobile-Icon} & \multicolumn{2}{c}{Pc-Text} & \multicolumn{2}{c}{Pc-Icon} & \multicolumn{2}{c}{Web-Text} & \multicolumn{2}{c}{Web-Icon} & \multicolumn{2}{c}{\textbf{Avg}} \\
\cmidrule(lr){2-3} \cmidrule(lr){4-5} \cmidrule(lr){6-7} \cmidrule(lr){8-9} \cmidrule(lr){10-11} \cmidrule(lr){12-13} \cmidrule(lr){14-15}
Method & v1 & v2 & v1 & v2 & v1 & v2 & v1 & v2 & v1 & v2 & v1 & v2 & v1 & v2 \\ 
\midrule
GUI-Actor-7B~\cite{webagent:guiact} & \textbf{94.9} & \textbf{96.5} & 82.1 & 84.3 & 91.8 & 91.7 & 80.0 & 84.1 & \textbf{91.3} & 93.9 & 85.4 & 82.3 & 88.3 & 89.5 \\
UI-TARS-72B~\cite{agent:uitars} & \textbf{94.9} & 94.8 & 82.5 & \textbf{86.3} & 89.7 & 91.2 & \textbf{88.6} & \textbf{87.9} & 88.7 & 91.5 & 85.0 & \textbf{87.7} & 88.4 & 90.3 \\
\midrule
MFT (Ours) & 93.4 & 95.7 & \textbf{82.8} & 85.9 & \textbf{96.4} & \textbf{97.8} & 84.3 & 87.3 & 90.8 & \textbf{94.3} & \textbf{89.4} & 84.4 & \textbf{89.2} & \textbf{91.3} \\
\bottomrule
\end{tabular}
}
\end{table*}

\subsection{Online Agentic RL with Hybrid Reward}
\begin{figure}[h!]
    \centering
    \includegraphics[width=0.95\linewidth]{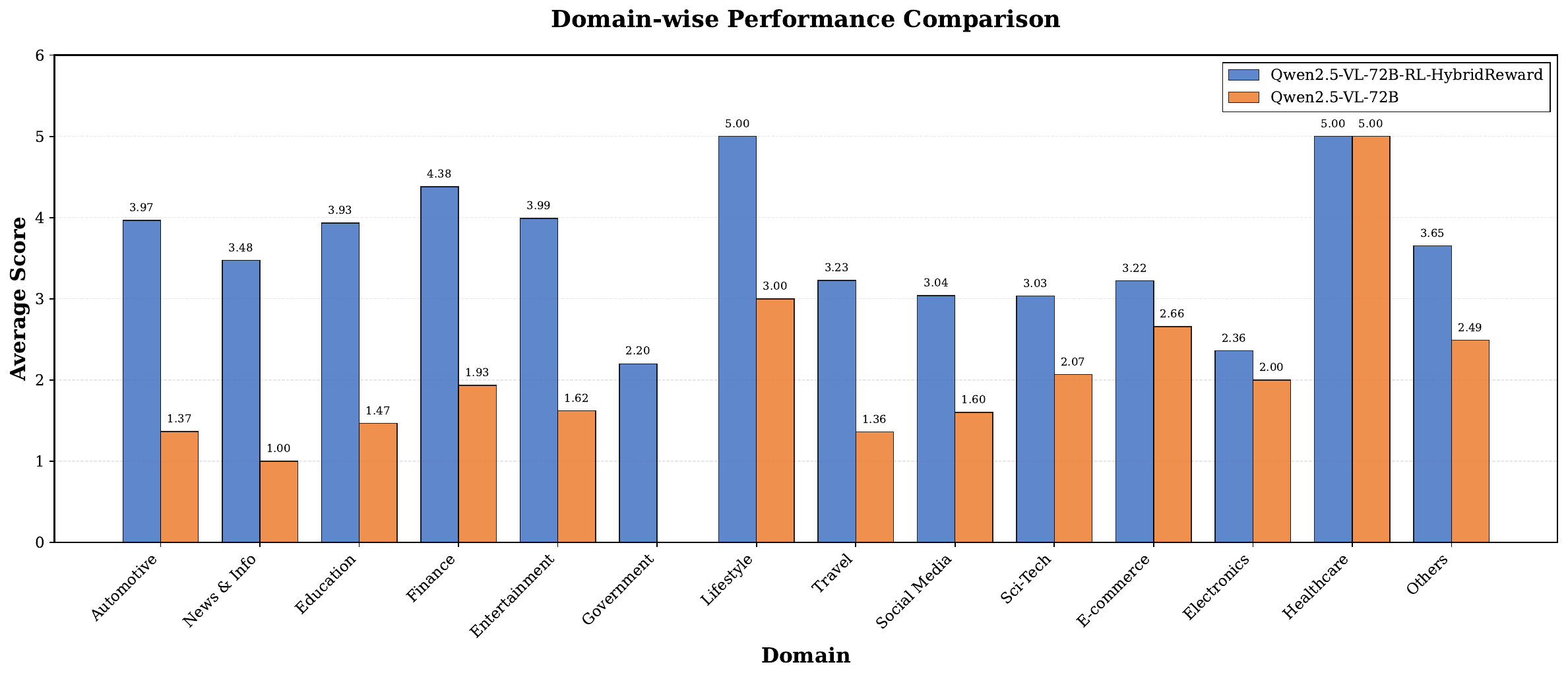}
    \caption{Detailed performance comparison across different sub-domains on Wild Websites. We compare the base Qwen2.5-VL-72B against the RL-tuned Qwen2.5-VL-72B-RL-HybridReward}
    \label{fig:rl_domain}
\end{figure}

To assess the generalization of the online RL-optimized models, we curated 87 
Wild Websites and user queries that were not included in the training 
distribution. We employ Webjudge as an automated evaluator to 
assign scores based on task completion and trajectory quality. As illustrated in Table~\ref{tab:webjudge_eval}, the model RL-optimized with Hybrid Reward achieves a substantial improvement of \textbf{1.55 points} 
in the average score compared to the baseline Qwen2.5-VL-72B. Notably, the model is trained through an iterative pipeline. Specifically, we harvest high-quality trajectories (those with high reward scores) generated during the Online RL phase and incorporate them into the Supervised Fine-Tuning (SFT) dataset. This iterative refinement significantly enhances the SFT model's capabilities, ultimately boosting the post-RL performance from 3.09 to 3.56.
As illustrated in Figure~\ref{fig:rl_domain}, the optimized model 
demonstrates a significant performance boost—exceeding \textbf{2 points}—across 
the majority of sub-domains, such as Automotive, News, Education, and Finance.

\begin{table}[h!]
\centering
\begin{threeparttable}
\caption{Performance comparison on Wild Websites. Zero means RL without SFT phase. The baseline is Qwen2.5-VL-72B-Instruct.}
\label{tab:webjudge_eval}
\begin{tabularx}{0.8\textwidth}{@{}Xccccc@{}}
\toprule
\textbf{Model} & \textbf{Max} & \textbf{Min} & \textbf{Avg\tnote{*}} & \textbf{Valid} & \textbf{InValid} \\ \midrule
Baseline & 5.0 & 1.0 & 2.01 & 75 & 12 \\ 
\begin{tabular}[c]{@{}l@{}}RL-HybridReward-Zero\end{tabular} & 5.0 & 0.0 & 3.09 & 70 & 17 \\ 
\begin{tabular}[c]{@{}l@{}}RL-HybridReward\end{tabular} & 5.0 & 1.0 & 3.56 & 77 & 10 \\ \bottomrule
\end{tabularx}

\begin{tablenotes}
    \footnotesize
    \item[*] Average score is calculated on a 5-point scale, excluding trajectories with negative scores, \ie, InValid Sample.
    \item \textbf{Valid}: Samples with score $\ge 1$; \textbf{InValid}: Samples with score of $-1$.
\end{tablenotes}
\end{threeparttable}
\end{table}

\begin{figure}[h!]
    \centering
    \includegraphics[width=0.8\linewidth]{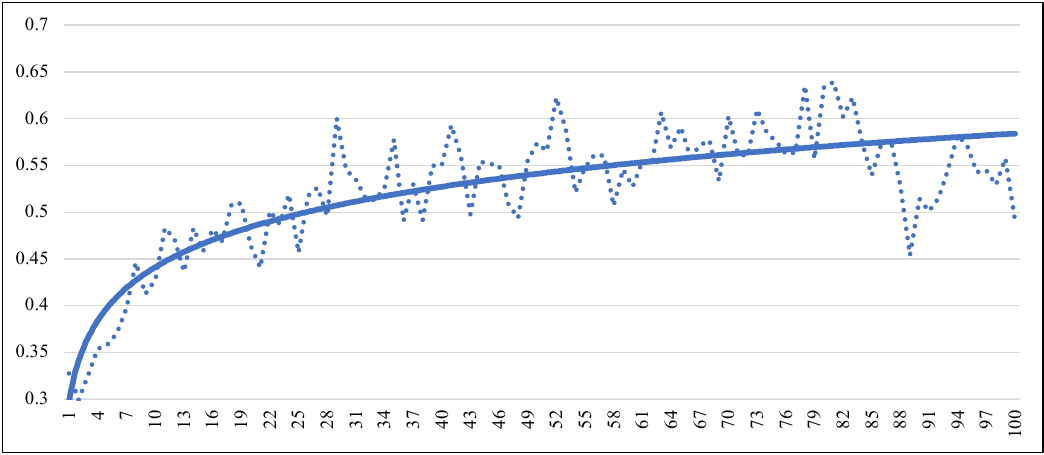}
    \caption{The training reward of Qwen3-VL-32B-Thinking-RL-HybridReward-Zero on WebArena.}
    \label{fig:webarena_reward}
\end{figure}

Beyond the evaluations on ``Wild Website'' datasets, we further validate the effectiveness of our \textit{HybridReward} method within a controlled and stable virtual environment. 
Specifically, we employ Qwen3-VL-32B-Thinking as the backbone model for online Reinforcement Learning (RL). 
To facilitate interaction during the RL rollout phase, we deployed a local instance of the WebArena environment on Alibaba Cloud ECS. 
Following the experimental setup of prior work~\cite{webmodel:tti}, we utilize queries from the test set for training; however, no ground-truth annotations are accessed during this process. 
Since our reward function is entirely \textbf{ground-truth-agnostic}, this approach can be characterized as a Test-Time Training (TTT) strategy. 
Notably, unlike existing methods~\cite{webmodel:tti} that often rely on hand-crafted \textit{tips} or site-specific instructions to guide the model, our training prompts are devoid of any domain-specific heuristics tailored to WebArena. 
This intentionally increases the complexity of the task and serves to underscore the intrinsic efficacy of our RL framework, demonstrating that the performance gains stem from the model's self-improvement rather than prompt engineering.
The overall training reward is illustrated in Figure~\ref{fig:webarena_reward}, which exhibits a consistent and healthy upward trend. 
For final evaluation, we report the performance at the 80th training step. 
During the inference phase, we maintain a rollout repetition of 5, \ie, Pass@5, consistent with the training configuration. 
The final results are summarized in Table~\ref{tab:rl_webarena}. 
As shown, our method achieves a $12.0\%$ absolute improvement over TTI---which employs a similar TTT strategy---and an $10.7\%$ gain over the vanilla Qwen3-VL-Thinking baseline.

\begin{figure}[h!]
    \centering
    \includegraphics[width=0.8\linewidth]{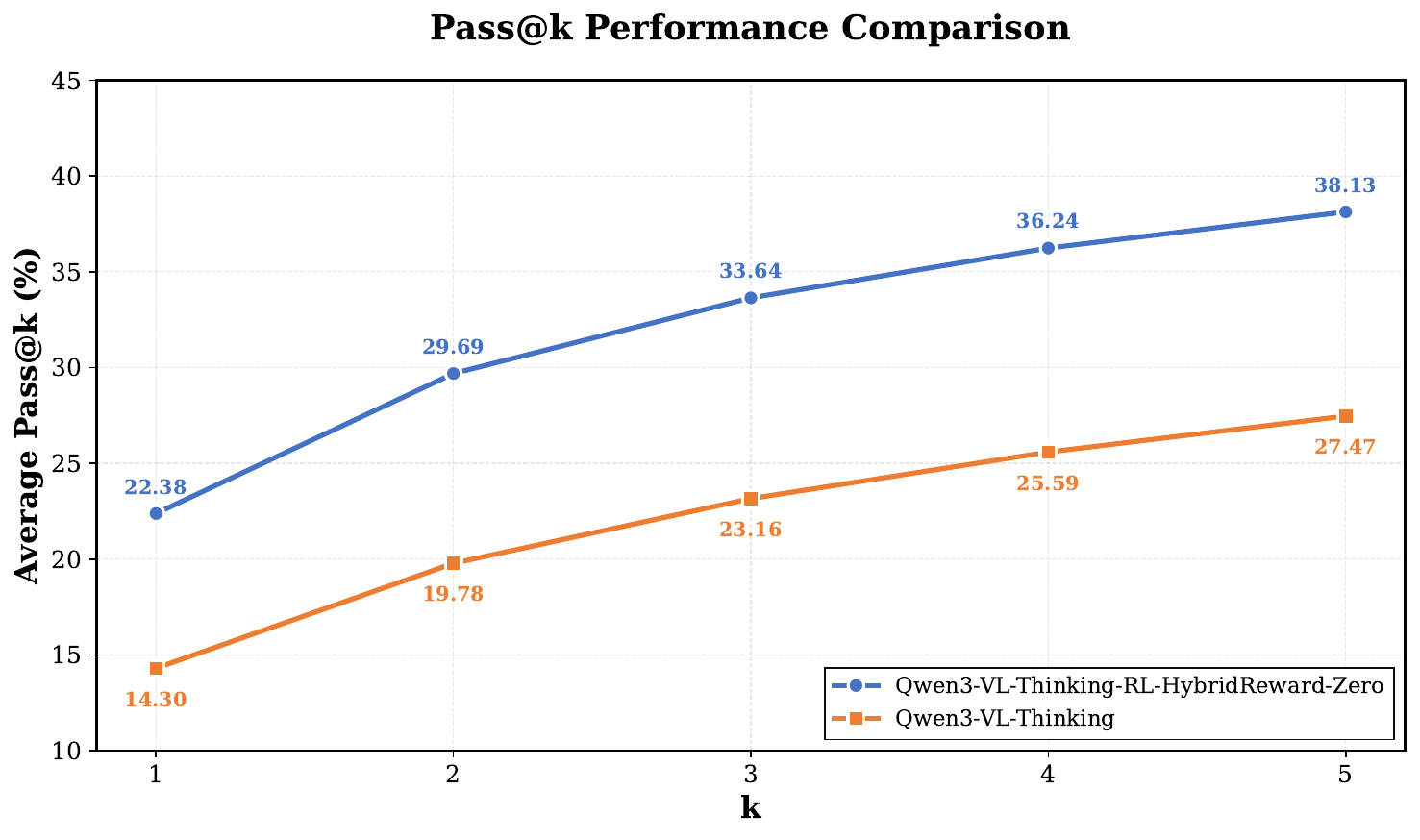}
    \caption{Performance comparison across different Pass@K on WebArena. We compare the base Qwen3-VL-32B-Thinking against the RL-tuned Qwen3-VL-32B-Thinking-RL-HybridReward-Zero.}
    \label{fig:pass_k_comparison}
\end{figure}

Furthermore, we compare the performance of the vanilla Qwen3-VL-Thinking model and its RL-enhanced counterpart across different $Pass@K$ metrics, as illustrated in Figure~\ref{fig:pass_k_comparison}. 
At $Pass@1$, our RL-trained model outperforms the baseline by $8.08\%$. 
Notably, as $K$ increases to $5$, the performance gap widens to $10.66\%$ ($Pass@5$). 
This increasing margin further demonstrates that the reinforcement learning process significantly reinforces the robustness of the model's predictions, enabling the agent to consistently identify successful execution paths within the expanded search space.

\begin{table}[h!]
\centering
\caption{Single model performance comparison across different web sites on WebArena. Our method demonstrates superior capabilities under the Test-Time Training setting.}
\label{tab:rl_webarena}
\resizebox{0.95\textwidth}{!}{
\begin{tabular}{lccccccc} 
\toprule
Method & Backbone & Overall & Shopping & CMS & Reddit & GitLab & Maps \\ \midrule

NNetnav~\cite{webmodel:nnetnav}  & Llama 3.1 8B & 7.2 & 7.4 & 4.2 & 0 & 0 & 28.5 \\
AutoWebGLM~\cite{webmodel:autowebglm}  & ChatGLM3 6B & 18.2 & - & - & - & - & - \\
AgentTrek~\cite{webmodel:agenttrek}  & Qwen2.5 32B & 22.4 & - & - & - & - & - \\
Learn-by-Interact~\cite{webmodel:learn_int}  & Codestral 22B & 24.2 & - & - & - & - & - \\
TTI~\cite{webmodel:tti} & Gemma 3 12B & 26.1 & 33.9 & 15.5 & 35.3 & 15.7 & 40.5 \\ \midrule
Baseline & Qwen3-VL-32B-Thinking & 27.4 & 30.7 & 34.1 & 28.3 & 30.0 & 8.2 \\
RL-HybridReward-Zero & Qwen3-VL-32B-Thinking & \textbf{38.1} & \textbf{40.0} & \textbf{37.6} & \textbf{36.1} & \textbf{35.4} & \textbf{42.3} \\ 
\bottomrule
\end{tabular}
}
\end{table}

\subsection{OpAgent Architecture for Web}
As previously demonstrated, relying solely on a standalone post-trained model is insufficient to achieve satisfactory performance in complex web agent tasks. To address this, we propose an agentic framework comprising four specialized components: a \textit{Planner}, a \textit{Grounder}, a \textit{Reflector}, and a \textit{Summarizer}. Specifically, we employ our supervised fine-tuned (SFT) Qwen2.5-VL-72B-MFT as the \textit{Grounder}. For the remaining sub-modules evaluated on WebArena, we utilize Gemini-3-Pro as the underlying backend. Consistent with prior studies~\cite{webagent:colorbrowser,webagent:ibmcuga}, we adopted a Human-in-the-Loop (HITL) strategy to incorporate site-specific tips for WebArena. Specifically, a tailored set of guidelines is curated for each individual website to provide the agent with necessary contextual hints. Ultimately, through a combination of refined prompt engineering and the seamless collaboration among our specialized agents, our framework attained the top position on the WebArena leaderboard, achieving a record-breaking score of $71.6\%$, as shown in Table~\ref{tab:agentic_webarena}.

\begin{table}[h!]
\centering
\caption{Agentic Architecture performance comparison across different web sites on WebArena.}
\label{tab:agentic_webarena}
\resizebox{0.95\textwidth}{!}{
\begin{tabular}{lccccccc} 
\toprule
Method & Backbone & Overall & Shopping & CMS & Reddit & GitLab & Maps \\ \midrule
SteP~\cite{webagent:Step} & - & 33.5 & - & - & - & - & - \\
Agent Workflow Memory~\cite{webagent:workflow_mem} & gpt-4 & 35.5 & 30.8 & 29.1 & 50.9 & 31.8 & 43.3 \\
GUI-API Hybrid Agent~\cite{webagent:gui_api} & gpt-4o & 35.8 & 34.6 & 26.4 & 50.9 & 36.7 & 46.8 \\
WebPilot~\cite{webagent:webpilot} & gpt-4o & 37.2 & 36.9 & 24.7 & 65.1 & 39.4 & 33.9 \\
AgentOccam-Judge~\cite{webagent:agentocaam} & gpt-4-turbo & 45.7 & 46.2 & 38.9 & 67.0 & 43.3 & 52.3 \\
Learn-by-Interact~\cite{webagent:learn_interact} & Codestral 22B & 48.0 & - & - & - & - & - \\
AgentSymbiotic~\cite{webagent:symbiotic} & claude-3.5-sonnet & 52.1 & 48.0 & 49.0 & 66.0 & 51.0 & 60.0 \\
ScribeAgent~\cite{webagent:scribeagent} & gpt-4o & 53.0 & 45.8 & 37.9 & 73.7 & 59.7 & 56.3 \\
WebOperator~\cite{webagent:weboperator} & gpt-4o & 54.6 & 49.2 & 55.0 & 76.4 & 52.8 & 55.2 \\
Jace.AI & - & 57.1 & - & - & - & - & - \\
OpenAI Operator & - & 58.1 & - & - & - & - & - \\
IBM CUGA~\cite{webagent:ibmcuga} & - & 61.7 & 58.3 & 62.6 & 75.5 & 61.7 & 64.2 \\
Narada AI & - & 64.2 & 57.2 & 63.2 & 74.5 & 73.9 & 58.7 \\
DeepSky Agent & - & 66.9 & - & - & - & - & - \\
GBOX AI & Claude Code & 68.0 & - & - & - & - & - \\ 
ColorBrowserAgent~\cite{webagent:colorbrowser} & GPT-5 & 71.2 & \textbf{72.9} & \textbf{76.4} & \textbf{87.4} & 65.7 & 55.9 \\ \midrule
\textbf{OpAgent (Ours)} & Gemini-3-Pro+Qwen2.5VL-MFT & \textbf{71.6} & 59.2 & 71.3 & 86.0 & \textbf{75.9} & \textbf{71.4} \\ 
\bottomrule
\end{tabular}
}
\end{table}

\section{Conclusion}
In this work, we focused on the task of Web Automation through a series of integrated research efforts, including multi-task supervised fine-tuning (SFT), online Agentic Reinforcement Learning (RL) in the wild, and the construction of a modular OpAgent Architecture. 
By leveraging the collaborative synergy of multiple specialized agents, we achieved a record-breaking success rate of 71.6\% on the WebArena benchmark, securing the top position on the leaderboard. 
However, the current paradigm remains heavily dependent on extensive prompt engineering and the complex orchestration of multiple agents, which incurs substantial human labor and computational overhead. 
Consequently, our future research will be directed toward enhancing the intrinsic exploration capabilities of individual models to reduce this reliance. Achieving this goal remains a formidable challenge.

\bibliographystyle{colm2024_conference}
\bibliography{custom}

\appendix
\clearpage

\end{document}